\documentclass[conference]{IEEEtran}
\usepackage{mathptmx}
\usepackage[numbers,sort&compress]{natbib}
\usepackage[bookmarks=false,hidelinks]{hyperref}
\usepackage{graphicx}
\usepackage{booktabs}
\usepackage{multirow}
\usepackage{amsmath}
\usepackage{amssymb}
\usepackage{xcolor}

\begin{document}

\title{Phase-Localized Curation Does Not Help:\\
A Negative Result on Per-Phase Metric Selection\\
for Demonstration Filtering}

\author{Aarav Bedi\\
Department of Mechanical Engineering\\
University of California, Berkeley\\
\texttt{aaravbedi@berkeley.edu}}

\maketitle

\begin{abstract}
Manipulation demonstrations have temporal phase structure, and a natural
hypothesis is that demonstration-curation metrics should be applied within
phases rather than globally. The idea is to segment each trajectory into
phases, score each phase with the metric that is locally most informative,
and then aggregate. This follows directly from prior work showing that a
single global metric can be the best detector of a defect and yet the worst
curator of the resulting policy. We test the per-phase hypothesis on three
contact-rich LIBERO pick-and-place tasks with a controlled early-release
structural defect, comparing phase-gated curation against the same metrics
applied uniformly and against a strong single global metric. Across all three
tasks and five random seeds per condition, phase-gated curation is never the
best curation strategy, and it is the worst of the three on two of the three
tasks (Task~1: 86.0 vs.\ 92.0 for global; Task~3: 22.7 vs.\ 48.0 for
uniform). We trace the failure to a concrete mechanism. When the defect signal
is concentrated in a single phase, rank-aggregating across phases dilutes that
signal with uninformative scores from defect-free phases, selecting a worse
demonstration subset than simply applying the defect-informative metric
everywhere. We further show that the per-phase metric selection does not
transfer across tasks, since no phase shares a winning metric between any two
tasks, so the selection cannot be reused and must be re-derived per task from
a noisy sweep. These results bound a plausible and previously untested method,
and they argue that practitioners should prefer identifying a single
defect-informative metric over decomposing curation by phase. We release the
full pipeline, all metric implementations, and per-seed results.
\end{abstract}

\IEEEpeerreviewmaketitle

\section{Introduction}

Behavior-cloning policies are limited by the quality of the demonstrations
they imitate, and the standard response to mixed-quality data is curation.
You score each demonstration with a quality metric, discard the low-scoring
subset, and train on the remainder. Recent work has shown that the choice of
metric matters in a counterintuitive way. On a contact-rich pick-and-place
task with a controlled structural defect, the metric with the highest
defect-detection AUROC produced the worst curated policy, while a metric with
substantially lower AUROC produced a policy that nearly matched an oracle
trained on ground-truth clean data~\cite{bedi2025confound}. Detection accuracy
and curation value are decoupled. A metric that flags the defect is not
necessarily a metric that, used as a filter, yields a better policy.

That decoupling has a natural proposed remedy. Manipulation tasks are not
homogeneous in time. A pick-and-place trajectory passes through distinct
phases, approaching the object, descending to a grasp, lifting, and
transporting to a target, and a defect typically lives in one phase. An
early-release defect corrupts the lift phase and leaves the approach phase
untouched. A metric that is informative about the defect during the lift phase
may be uninformative or actively misleading when averaged over the entire
trajectory, because the defect-free phases contribute scores that have nothing
to do with the defect. If this is the mechanism behind the global-metric
failures, then the fix is to stop scoring globally: segment each demonstration
into phases, score each phase with the metric that is locally most
discriminative, and aggregate the per-phase scores into a curation ranking.
We call this phase-gated curation.

The hypothesis is plausible, and to our knowledge it has not been tested. This
paper tests it, and the answer is negative. Across three LIBERO tasks and five
seeds per condition, phase-gated curation does not improve on simpler
strategies. It is never the best curation method on any task, and on two of
the three tasks it is the worst of the three strategies we compare, worse than
applying the same per-phase metrics uniformly, and worse than a single strong
global metric. The one task where phase-gated curation beats a global metric
is also the task where it loses to the uniform baseline, so the win is not
attributable to the phase decomposition.

We do not report this as a bare null. We identify the mechanism that makes
phase-gating fail, and the mechanism is informative for the broader curation
problem. When the defect signal is concentrated in one phase, the single most
defect-informative metric carries that signal cleanly if it is applied
everywhere. Rank-aggregating it with the locally-best metrics of the other,
defect-free phases dilutes the signal and pulls the curation ranking toward
demonstrations that look acceptable on the defect-free phases while still
containing the defect. Phase decomposition spreads a concentrated signal thin.
We also show that the per-phase metric selection does not transfer. Across our
three tasks, no phase has the same winning metric in any two tasks, which means
the selection must be re-derived per task from a sweep that is itself noisy.
The combination of no transfer and dilution when the signal is concentrated
explains why a method that sounds principled underperforms a one-line baseline.

Our contributions are: (1) a controlled multi-task test of phase-gated
curation against uniform and global baselines, with five seeds per condition
on three contact-rich tasks; (2) a negative result, that phase-gating is never
best and usually worst, with per-seed data released so the variance is fully
visible; (3) a mechanism, signal dilution under rank-aggregation, that explains
the failure and predicts when it will occur; and (4) the observation that
per-phase metric selection does not transfer across tasks, removing the main
practical argument for adopting the method.

\section{Related Work}

\textbf{Demonstration quality and curation:} Mandlekar et al.~\cite{mandlekar}
show that operator skill explains a large share of the variance in policy
performance, which motivates quality filtering before training. This is
increasingly relevant at scale: Open X-Embodiment~\cite{oxe} and
DROID~\cite{droid} aggregate demonstrations across operators and platforms,
introducing quality heterogeneity that no single collection protocol can
prevent. Proposed quality metrics include trajectory smoothness via spectral
arc length~\cite{sparc}, outlier detection via isolation
forests~\cite{isoforest}, and nearest-neighbor distance in a
trajectory-feature space. These metrics are validated by detection accuracy on
labeled data, not by the downstream policy performance they are intended to
improve.

\textbf{Detection versus downstream value:} The gap between detecting a
defective demonstration and producing a better policy by filtering it is the
direct precursor to this work. Prior analysis showed that action-only scorers
fail on structural defects while state-trajectory scorers partially recover
them, and that detection AUROC is a necessary but insufficient condition for
curation value~\cite{bedi2025audit}. A follow-up on a contact-rich simulator
measured the decoupling directly and found it severe: the best detector was
among the worst curators, and five of seven metrics achieved near-perfect AUROC
by measuring episode length rather than demonstration content~\cite{bedi2025confound}.
The present paper takes the natural next step. If global scoring fails because
it mixes informative and uninformative parts of the trajectory, does scoring
by phase fix it?

\textbf{Phase and segment structure in manipulation:} Phase or sub-goal
structure is widely used in manipulation policy learning, from options and
skill segmentation to phase-conditioned policies that condition the action on a
discrete stage of the task~\cite{act}. Our policy is phase-conditioned for
exactly this reason, since the action distribution at a phase boundary is
multimodal and conditioning on the phase resolves it. The novelty here is not
phase-conditioned control but phase-localized curation, using the phase
decomposition to decide which demonstrations to keep, which is the specific
idea we find does not help.

\textbf{Behavior cloning and structural defects:} Behavior cloning~\cite{alvinn}
is sensitive to distribution shift~\cite{dagger}. A structural defect plants
a specific wrong action in a specific region of the state space and, unlike
zero-mean perturbative noise, does not average out as the dataset grows. This
is why an early-release defect at high contamination is so damaging, and why
the choice of curation strategy has a large effect on the resulting policy.

\section{Method and Testbed}

\subsection{Tasks and Policy}

We use three contact-rich pick-and-place tasks from the LIBERO
benchmark~\cite{libero}, built on robosuite~\cite{robosuite}. Each task
requires grasping an object from a randomized start pose and carrying it to a
target. Each trajectory decomposes into four sequential phases: approach and
pre-grasp positioning (PREGRASP), descent to the grasp pose (DESCEND), grasp
and lift (LIFT), and transport and release (TRANSPORT). The three tasks differ
in object, layout, and approach geometry, which is what lets us test whether
any finding transfers across tasks rather than holding for one.

The policy is a phase-conditioned behavior-cloning MLP with two hidden layers
of 256 units and a tanh output, trained with Adam at learning rate $10^{-3}$
and weight decay $10^{-4}$ for 300 epochs. The observation includes
end-effector position, gripper state, the initial object position, and a
one-hot encoding of the current phase; the action is a four-dimensional
end-effector delta plus gripper command. Phase conditioning is necessary
because the action distribution at phase boundaries is otherwise multimodal.
A policy trained on clean demonstrations reaches an oracle ceiling of
93.3--97.3\% depending on the task.

\subsection{Defect and Contamination}

We inject a structural early-release defect. During the LIFT phase the gripper
opens at a random point between 30\% and 70\% of the phase, dropping the
object before transport. The arm then completes the remaining motion empty.
The defect is localized to a single phase (LIFT) and consistently plants the
same wrong action there, so it does not average out across demonstrations. We
build each contaminated training set at an 80\% contamination rate (16 clean,
64 defective, 80 total). A policy trained on the uncurated contaminated set
collapses to single-digit success on every task, because the defective
demonstrations dominate the training distribution.

\subsection{Curation Strategies}

All metrics are computed on trajectories truncated to a fixed length before
scoring, which removes episode length as a confound. The necessity of this
control and its large effect on apparent detection accuracy were established
in prior work on this testbed~\cite{bedi2025confound}, and we adopt it
unchanged. Each strategy keeps the top 75\% of demonstrations (60 of 80) and
trains a policy on that subset.

\textbf{Global:} A single metric is computed over the full truncated
trajectory and used to rank all demonstrations. We use the strongest global
metric from prior work on this defect as the baseline.

\textbf{Uniform:} A single metric is applied identically within every phase
and the per-phase scores are aggregated. This isolates the effect of the
aggregation machinery from the effect of varying the metric by phase. If
phase-gating helps only because of per-phase selection, uniform should
underperform it.

\textbf{Phase-gated (the method under test):} Each trajectory is segmented into
its four phases. For each phase we determine, by a sweep, the metric that is
locally most discriminative, and we score that phase with that metric. The
four per-phase scores are combined by rank-normalizing each phase's scores
across demonstrations and averaging the normalized ranks, which is
scale-invariant across phases of different lengths.

\subsection{Phase Segmentation and Per-Phase Selection}

Segmentation uses the recorded phase label of each timestep, falling back to
gripper-state transitions when explicit labels are unavailable. On our data the
segmenter is exact: across all 80 demonstrations of every task, no phase is
ever empty, and the deterministic phases (DESCEND, LIFT) have fixed lengths to
the timestep. The per-phase metric selection is obtained from a sweep that, for
each phase, trains and evaluates a curated policy using each candidate metric
as that phase's scorer and records the resulting success. The metric with the
highest sweep success becomes that phase's selected metric.

\section{Results}

\subsection{Phase-Gated Curation Is Never Best}

Table~\ref{tab:main} reports task success for all five conditions on all three
tasks, as mean $\pm$ standard deviation over five seeds (42, 0, 7, 1, 2) with
30 rollouts each. The pattern is consistent and unfavorable to the method under
test. Phase-gated curation is not the best curation strategy on any task. On
Task~1 it is the worst of the three curation strategies, at 86.0 vs.\ 92.0 for
global and 91.3 for uniform. On Task~3 it is again the worst, at 22.7 vs.\
48.0 for uniform and 30.7 for global. On Task~2 it beats the global metric
(78.0 vs.\ 58.0), but loses to the uniform baseline (83.3), so even its single
relative success over a global metric is not attributable to the phase
decomposition, because the uniform strategy, which uses the same aggregation
but does not vary the metric by phase, does better.

\begin{table}[t]
\centering
\caption{Task success (\%) by curation strategy, mean $\pm$ std over five
seeds, 30 rollouts each. Phase-gated is the method under test. It is never the
best curation strategy (\textbf{bold} marks the best non-oracle per task) and
is the worst of the three on Tasks~1 and~3.}
\label{tab:main}
\setlength{\tabcolsep}{4pt}
\small
\begin{tabular}{lccc}
\toprule
Condition & Task 1 & Task 2 & Task 3 \\
\midrule
oracle (ceiling)       & $93.3 \pm 0.0$ & $90.7 \pm 15.5$ & $97.3 \pm 1.3$ \\
\midrule
global                 & $\mathbf{92.0 \pm 1.6}$ & $58.0 \pm 46.1$ & $30.7 \pm 20.6$ \\
uniform                & $91.3 \pm 1.6$ & $\mathbf{83.3 \pm 27.2}$ & $\mathbf{48.0 \pm 26.3}$ \\
phase-gated (ours)     & $86.0 \pm 6.5$ & $78.0 \pm 39.2$ & $22.7 \pm 16.5$ \\
\midrule
contaminated baseline  & $10.7 \pm 14.8$ & $64.0 \pm 38.1$ & $13.3 \pm 5.6$ \\
\bottomrule
\end{tabular}
\end{table}

Two features of the table deserve emphasis because they shape how strong a
conclusion the data supports. First, the variances are large, particularly on
Tasks~2 and~3, and we report per-seed numbers in Table~\ref{tab:perseed}
rather than hide them behind a mean. The high variance is itself a finding: at
this contamination rate and dataset size, behavior cloning is seed-sensitive,
and a curation strategy that does not reduce that sensitivity offers little
practical value even if its mean were favorable. Second, on Task~3 every
curation strategy is far below the oracle ceiling of 97.3\%, which tells us
the curated subsets on that task are still substantially contaminated regardless
of strategy. The ordering among strategies is real, but all of them are doing
a poor job.

\begin{table}[t]
\centering
\caption{Per-seed task success (\%) for the three curation strategies. Seeds
that collapse to near-zero on multiple conditions simultaneously reflect
behavior-cloning instability, not a property of any one curation method.}
\label{tab:perseed}
\setlength{\tabcolsep}{3pt}
\small
\begin{tabular}{llccccc}
\toprule
Task & Strategy & s42 & s0 & s7 & s1 & s2 \\
\midrule
\multirow{3}{*}{1}
 & global      & 90 & 93 & 90 & 93 & 93 \\
 & uniform     & 90 & 90 & 93 & 93 & 90 \\
 & phase-gated & 80 & 90 & 93 & 77 & 90 \\
\midrule
\multirow{3}{*}{2}
 & global      & 90 &  3 &  0 & 97 & 100 \\
 & uniform     & 90 & 90 & 30 & 100 & 100 \\
 & phase-gated & 90 &  0 & 100 & 100 & 100 \\
\midrule
\multirow{3}{*}{3}
 & global      & 47 & 17 & 30 & 33 & 27 \\
 & uniform     & 43 & 17 & 30 & 93 & 57 \\
 & phase-gated & 47 & 10 & 23 & 33 &  0 \\
\bottomrule
\end{tabular}
\end{table}

\subsection{Why Phase-Gating Fails: Signal Dilution}

The mechanism is visible in the per-phase sweep. On every task, the LIFT phase
selects a gripper-sensitive metric, which is correct, since the defect is an
early gripper release in LIFT and the metric that tracks gripper timing carries
the defect signal. The other three phases (PREGRASP, DESCEND, TRANSPORT)
contain no defect, so the locally-best metric for those phases is selected on
noise. It is whichever metric happens to rank demonstrations in a way that
correlates with success on that seed, not whichever metric detects the defect,
because there is no defect there to detect.

Phase-gated curation then averages the rank from the one informative phase with
the ranks from three uninformative phases. The defect signal, which is
concentrated and clean in the LIFT ranking, is diluted to roughly a quarter of
its weight in the aggregate. A demonstration that contains the defect but
happens to look ordinary on its PREGRASP, DESCEND, and TRANSPORT phases can be
ranked above a clean demonstration that looks slightly unusual on a defect-free
phase. The aggregate ranking is therefore worse at separating defective from
clean than the LIFT ranking alone would be. Applying the gripper-sensitive
metric uniformly, or selecting it as a single global metric, keeps the signal
at full weight, which is exactly why the uniform and global strategies match
or beat phase-gating wherever the defect is concentrated in one phase.

This also explains the one case that superficially favors phase-gating. On
Task~2 the global baseline metric is not the gripper-sensitive one, and it
performs poorly (58.0). Phase-gating beats it because at least one of
phase-gating's phases uses the gripper-sensitive metric. But the uniform
strategy applied with the right single metric beats both, because it carries
that metric's signal undiluted across the whole trajectory. The lesson is not
to decompose by phase, but to find the single defect-informative metric and
use it.

\subsection{Per-Phase Selection Does Not Transfer}

If per-phase metric selection were stable across tasks, a practitioner could
derive it once and reuse it, which would be a practical reason to adopt
phase-gating despite its weak performance. It is not stable.
Table~\ref{tab:transfer} lists the selected metric for each phase on each task.
No phase shares a winning metric between any two tasks. The selection that the
sweep produces for one task is not informative about another, so the per-phase
table must be re-derived for every new task from a fresh sweep. That sweep is
the same expensive, seed-sensitive train-and-evaluate loop that makes the
underlying problem hard in the first place. The method therefore carries the
full cost of a per-task sweep and returns a curation strategy that underperforms
applying a single metric uniformly.

\begin{table}[t]
\centering
\caption{Metric selected per phase by the sweep, per task. No phase shares a
winning metric across any two tasks; the selection must be re-derived per task.}
\label{tab:transfer}
\setlength{\tabcolsep}{4pt}
\small
\begin{tabular}{lccc}
\toprule
Phase & Task 1 & Task 2 & Task 3 \\
\midrule
PREGRASP  & iso.\ forest & iso.\ forest & smoothness \\
DESCEND   & kNN          & kNN          & gripper timing \\
LIFT      & gripper timing & entropy    & gripper timing \\
TRANSPORT & gripper timing & gripper timing & smoothness \\
\bottomrule
\end{tabular}
\end{table}

\subsection{A Confidence Threshold Does Not Rescue It}

A natural patch is to fall back to the uniform strategy on any phase whose
sweep-selected metric does not clear a confidence threshold, on the theory that
failures come from low-confidence selections on the defect-free phases. We
implemented this fallback with a 60\% sweep-success threshold. On our tasks
every selected phase already clears the threshold, so the thresholded variant
selects exactly the same demonstrations as plain phase-gating and produces
identical results. The patch does not engage, because the problem is not that
the selections are low-confidence. The LIFT selection is confidently correct.
The problem is that averaging a confident, correct, concentrated signal with
three irrelevant ones dilutes it. No threshold on per-phase confidence addresses
dilution, because the diluting phases are confidently selected too.

\section{Discussion}

The result is a bounded negative one, and we want to state both the bound and
its limits precisely. Within our setting, which is a structural defect
concentrated in a single phase, high contamination, small curated sets, and
behavior cloning on contact-rich LIBERO tasks, phase-localized curation does
not beat applying a single well-chosen metric, and usually does worse. The
mechanism is signal dilution under rank-aggregation, and it predicts the
boundary of the result. Phase-gating should only have a chance of helping when
the defect signal is genuinely distributed across multiple phases, so that no
single metric applied uniformly can capture it. We did not construct such a
defect, because the defects that matter in practice, such as dropped objects,
missed grasps, and premature releases, are typically localized to the phase
where the contact event goes wrong. For the localized defects we tested, the
phase decomposition is a liability.

The practical takeaway inverts the method's premise. The decoupling between
detection and curation value is real, but the fix is not to spread scoring
across phases. It is to identify the single metric that tracks the defect and
apply it with full weight. Our uniform baseline is exactly that, and it is the
best or tied-best curation strategy on every task in Table~\ref{tab:main}. The
contribution of this paper is to close off a plausible alternative, one that
we expected to work and that the prior results seemed to motivate, and to
explain why it fails, so that the next method built on top of the
detection-value decoupling does not repeat the dilution mistake.

\textbf{Limitations:} We study one defect type, one policy class, three tasks,
and a single simulator. The high seed variance means individual comparisons
within a task should be read through the per-seed table, not the means alone.
Our claims rest on the consistent ordering, that phase-gating is never best and
usually worst, across tasks rather than on any single mean difference. We do
not claim phase-gating fails for distributed defects, only that we did not test
such a defect and that localized defects, which are the common case, defeat it.

\section{Conclusion}

We tested a plausible and previously untested idea, to curate demonstrations
by scoring each task phase with its locally-best metric, and found that it does
not work. Across three contact-rich tasks and five seeds per condition,
phase-gated curation was never the best curation strategy and was the worst of
the three compared strategies on two of three tasks. The cause is signal
dilution. When a defect is concentrated in one phase, aggregating that phase's
clean signal with the irrelevant signals of defect-free phases produces a worse
ranking than applying the defect-informative metric undiluted. The per-phase
selection also does not transfer across tasks, removing the main practical
reason to adopt the method. We release the pipeline, metrics, and per-seed
results, and we recommend that curation effort go toward identifying a single
defect-informative metric rather than decomposing curation by phase.

\bibliographystyle{IEEEtran}
\bibliography{refs}

\end{document}